\documentclass[letterpaper, 10 pt, conference]{ieeeconf}  %

\IEEEoverridecommandlockouts                              %

\title{\LARGE \bf
Probabilistic Traversability Model for Risk-Aware Motion Planning in Off-Road Environments%
}

\author{Xiaoyi Cai$^1$, Michael Everett$^2$, Lakshay Sharma$^1$, 
Philip R. Osteen$^3$, and Jonathan P. How$^1$%
\thanks{$^1$Massachusetts Institute of Technology, Cambridge, MA 02139, USA. {\tt\{xyc, lakshays, jhow\}@mit.edu}.}%
\thanks{$^2$Northeastern University, Boston, MA 02115, USA. {\tt m.everett@northeastern.edu}.}%
\thanks{$^{3}$DEVCOM Army Research Laboratory, Adelphi, MD 20783, USA. {\tt philip.r.osteen.civ@army.mil}. Distribution Statement A. Approved for public release: distribution unlimited.}%
}

\newcommand\copyrighttext{%
  \scriptsize © 2023 IEEE. The paper will appear in the 2023 IEEE/RSJ International Conference on Intelligent Robots and Systems (IROS 2023). Personal use of this material is permitted.  Permission from IEEE must be obtained for all other uses, in any current or future media, including reprinting/republishing this material for advertising or promotional purposes, creating new collective works, for resale or redistribution to servers or lists, or reuse of any copyrighted component of this work in other works.
  }
\newcommand\copyrightnotice{%
\begin{tikzpicture}[remember picture,overlay]
\node[anchor=south,yshift=10pt] at (current page.south) {\fbox{\parbox{\dimexpr\textwidth-\fboxsep-\fboxrule\relax}{\copyrighttext}}};
\end{tikzpicture}%
}

\usepackage{amsmath,amsthm,amssymb, amsfonts} %
\usepackage{bbold}
\usepackage{mathtools}
\usepackage{graphicx, xcolor}
\usepackage{pdfpages}
\usepackage{bm}
\usepackage{url}
\usepackage{enumerate}
\usepackage{float,balance}
\usepackage[sort, compress]{cite}
\usepackage{cases,balance}
\usepackage[pdftex, pdfstartview={FitV}, pdfpagelayout={TwoColumnLeft},bookmarksopen=true,plainpages = false, colorlinks=true, linkcolor=black, citecolor = black, urlcolor = black,filecolor=black , pagebackref=false,hypertexnames=false, plainpages=false, pdfpagelabels ]{hyperref}

\usepackage[capitalise]{cleveref}

\usepackage[font=footnotesize]{caption}
\usepackage{subcaption}

\usepackage{array}
\usepackage{booktabs}
\usepackage{multirow} %

\usepackage{comment} %
\excludecomment{comments} %

\usepackage{algorithm}
\usepackage[noend]{algpseudocode} %
\definecolor{commentclr}{RGB}{110, 149, 204}

\makeatletter
\newcommand\fs@spaceruled{\def\@fs@cfont{\bfseries}\let\@fs@capt\floatc@ruled
  \def\@fs@pre{\vspace{0.6\baselineskip}\hrule height.8pt depth0pt \kern2pt}%
  \def\@fs@post{\kern2pt\hrule\relax}%
  \def\@fs@mid{\kern2pt\hrule\kern2pt}%
  \let\@fs@iftopcapt\iftrue}
\makeatother

\usepackage{tikz}

\newcommand{\setO}{\boldsymbol{O}} %

\newcommand{\Param}{\boldsymbol{\psi}}
\newcommand{\setParam}{\boldsymbol{\Psi}}

\newcommand{\setX}{\mathbf{X}} %

\newcommand{\setM}{\mathbf{M}} %

\newcommand{\btheta}{\bm{\theta} } %

\newcommand{\done}[1]{\mathbb{1}^{\text{done}}(#1)} %

\newcommand{\lcvar}[2]{{\text{CVaR}_{#1}^{\leftarrow}(#2)}}
\newcommand{\rcvar}[2]{{\text{CVaR}_{#1}^{\rightarrow}(#2)}}
\newcommand{\lvar}[2]{{\text{VaR}_{#1}^{\leftarrow}(#2)}}
\newcommand{\rvar}[2]{{\text{VaR}_{#1}^{\rightarrow}(#2)}}

\newcommand{\cvarcost}{{CVaR-Cost}}
\newcommand{\cvardyn}{{CVaR-Dyn}}

\newcommand\norm[1]{\left\lVert#1\right\rVert}

\newcommand{\R}{\mathbb{R}}

\newcommand{\tr}{^\top}
\newcommand*{\defeq}{:=}

\theoremstyle{plain}%

\theoremstyle{remark}
\newtheorem{remark}{Remark}

\theoremstyle{definition}

\graphicspath{{Figs/}}

\begin{document}

\maketitle
\thispagestyle{empty}
\pagestyle{empty}

\begin{abstract}
A key challenge in off-road navigation is that even visually similar terrains or ones from the same semantic class may have substantially different traction properties. Existing work typically assumes no wheel slip or uses the expected traction for motion planning, where the predicted trajectories provide a poor indication of the actual performance if the terrain traction has high uncertainty. 
In contrast, this work proposes to analyze terrain traversability with the empirical distribution of traction parameters in unicycle dynamics, which can be learned by a neural network in a self-supervised fashion. The probabilistic traction model leads to two risk-aware cost formulations that account for the worst-case expected cost and traction. To help the learned model generalize to unseen environment, terrains with features that lead to unreliable predictions are detected via a density estimator fit to the trained network's latent space and avoided via auxiliary penalties during planning. Simulation results demonstrate that the proposed approach outperforms existing work that assumes no slip or uses the expected traction in both navigation success rate and completion time. Furthermore, avoiding terrains with low density-based confidence score achieves up to 30\% improvement in success rate when the learned traction model is used in a novel environment.
\end{abstract}

{\small
\section*{Supplementary Material}
Video and GPU implementation of planners are available at \url{https://github.com/mit-acl/mppi_numba}.
}

\copyrightnotice

\section{Introduction}
Progress in autonomous robot navigation has expanded the set of non-urban environments where robots can be deployed, such as mines, forests, oceans and Mars~\cite{fan2021step,frey2022locomotion,pereira2013risk, massari2004autonomous}. Unlike environments where safe and reliable navigation can be achieved by avoiding hazards easily detected based on geometric features, navigation in forested environments poses unique challenges that still prevent systems from achieving good performance, because a purely geometric view of the world is not sufficient to identify non-geometric hazards (e.g., mud puddles, slippery surfaces) and geometric non-hazards (e.g., grass and foliage). To this end, recent approaches train semantic classifiers for camera images~\cite{valada16iser,valada2017adapnet, chen2022cali, guan2022ga, guantns} or lidar pointclouds~\cite{shaban2022semantic} to identify terrains or objects that could cause failures to the robotic platform in order to manually design semantics-based cost functions for navigation. However, existing labeled datasets for off-road navigation have limited number of class labels such as ``bush'', ``grass'', and ``tree'' for vegetation without capturing the varying traversability within each class (e.g.,~\cite{RUGD2019IROS,jiang2021rellis}), or have limited transferability due to specificity to the vehicles that collected the data.

\begin{figure}[t]
	\centering
	\includegraphics[width=\linewidth, trim={0cm 0cm 0cm 0cm},clip]{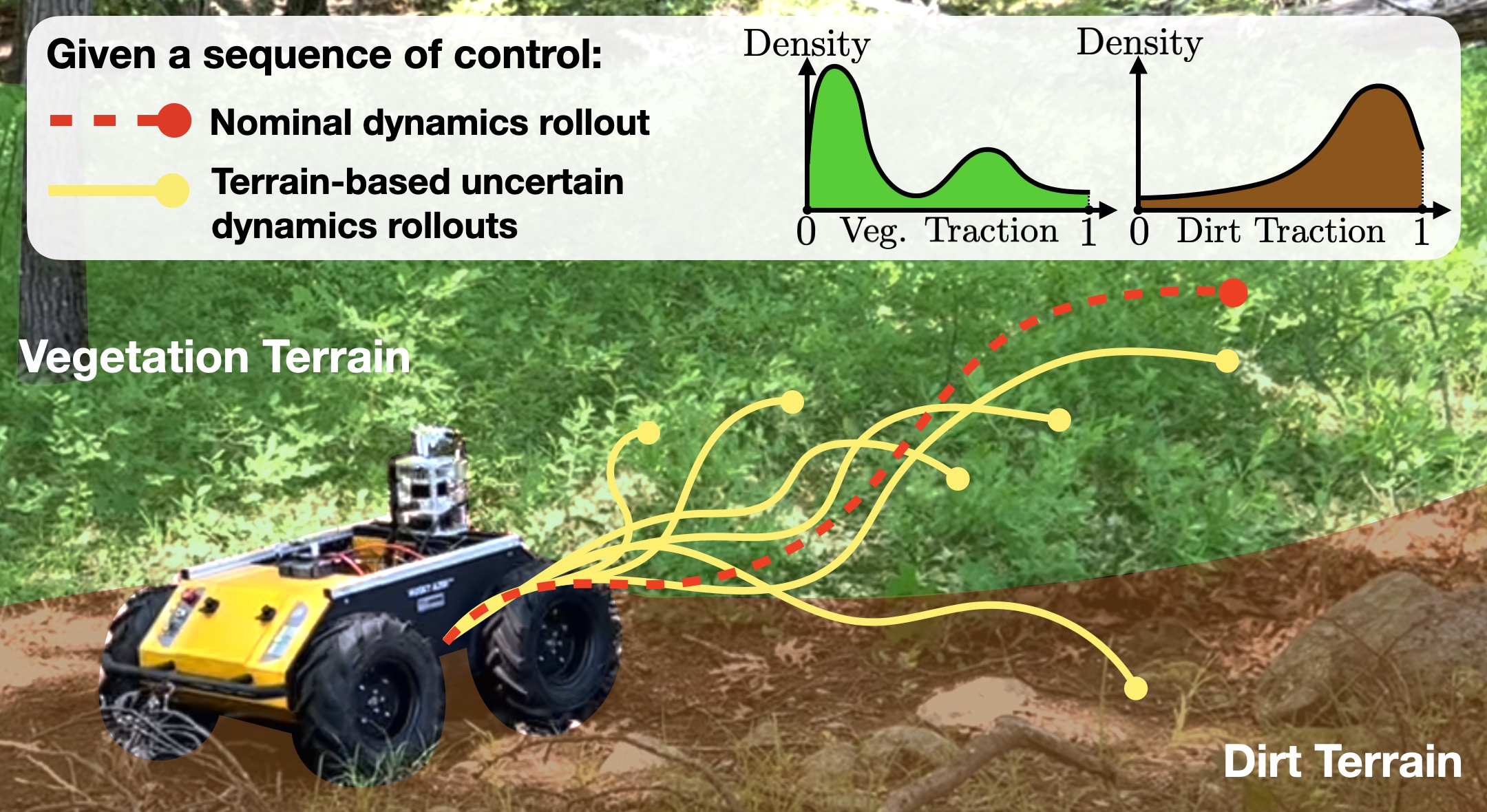}
	\caption{Imperfect sensing and coarse semantic labels about the terrain make it difficult to model traction and future states when given a control sequence. This work proposes to learn empirical traction distribution in order to estimate the worst-case expected outcomes for risk-aware planning.}
	\label{fig:highlevel_figure}
 \vspace*{-0.2in}
\end{figure}

While manually designing cost functions based on semantically labeled terrains are common, self-supervised learning techniques are increasingly adopted to model traversability based on historically collected data about terrain properties~\cite{kahn2021badgr, yao2022rca, zurn2020self}. 
For example, a robot can learn to map camera images to terrain properties such as traction that affects achievable velocities~\cite{Gasparino2022wayfast, cai2022risk, sathyamoorthy2022terrapn}.
However, these methods do not fully account for the uncertainty in the learned terrain models. Based on our empirical findings (see the real-world traction distribution learned by a neural network (NN) in the ``Traversability Analysis'' block of Fig.~\ref{fig:traversability_analysis_pipeline}), terrains properties such as traction can be non-Gaussian, which makes assuming no slip or using the expected traction inadequate to capture the risk of obtaining poor performance.
Even if the full empirical distribution can be learned about the traction parameters, working with distributions that are not necessarily Gaussian makes it difficult to efficiently characterise and estimate the cost of experiencing tail events during planning.

In this work, we propose to analyze terrain traversability with the empirical distribution of traction parameters in the unicycle model without making Gaussian assumptions. In addition, we propose two cost formulations that exploit the learned traction distribution to estimate the impact of tail events on navigation performance. The resultant planners mitigate the risks that stem from the stochastic dynamics and the downstream realization of the objectives. Lastly, as a NN's predictions are unreliable when input features are significantly different from training data, we fit a Gaussian Mixture Model (GMM) in the latent space of the trained NN to provide confidence scores for detecting and avoiding novel terrains via auxiliary planning costs. In summary, the contributions of this work are:
\begin{itemize}
    \item A new representation of traversability as an empirical traction distribution in the unicycle dynamics conditioned on both semantic and geometric terrain features;
    \item Two risk-aware cost formulations based on the worst-case expected \textit{traction} and \textit{objective} that outperform methods that assume no slip or use expected traction;
    \item A GMM-based detector for detecting terrains that may lead to unreliable NN predictions and should be avoided during planning, which improves navigation success rate by 30\% when a learned traction model is used in an environment unseen during training.
    
\end{itemize}
\section{Related Work}

\subsection{Traversability Representation}

Traversability analysis is a key component of off-road navigation algorithms; a more complete summary of various approaches is provided in~\cite{cai2022risk}, including representations based on proprioceptive measurements~\cite{oliveira2021three, otte2016recurrent}, geometric features~\cite{larson2011off, overbye2020fast, overbye2022g, fan2021step} and combination of geometric and semantic features~\cite{guan2021ttm, tan2021risk, shaban2022semantic}.
WayFast~\cite{Gasparino2022wayfast} is a more recent approach, similar to this work, that proposes to represent traversability by learning traction coefficients for a unicycle model from terrain perception data.
Another recent work~\cite{frey2022locomotion} represents traversability as the probability of a quadruped robot to stabilize itself on uneven terrain, based on 3D occupancy data of the terrain.
In~\cite{yang2022learning}, speed and gait policies are learned based on terrain semantics and human demonstrations; these policies provide a novel interpretation of the terrain's traversability and can be used by the robot's motor control policy.
A key limitation, however, is that these point estimates of traversability do not capture the uncertainty of terrain properties on similar-looking terrains. Instead, our approach represents traversability as the \textit{distribution of traction parameters} in the dynamics model.

\subsection{Planning with Terrain-Dependent Stochastic Dynamics}

After learning the traction distribution in the dynamics model, a further challenge exists in incorporating this model into the planner.
Despite capturing various types of uncertainties and risks, many existing methods still plan with the nominal or expected parameters in the dynamics model, such as~\cite{tan2021risk,fan2021learning, cai2022risk}.
Alternatively, our work is inspired by~\cite{wang2021adaptive} that proposes a general framework for optimizing the conditional value at risk (CVaR) of the objective under uncertain dynamics, parameters and initial conditions, by taking extra samples from sources of uncertainty. 
In this work, we evaluate each control sequence based on traction samples and use the CVaR of the noisy realizations of the objectives instead of the nominal objective.
Additionally, we propose a more computationally efficient approach which computes the nominal objective using trajectory rollouts based on the CVaR of traction parameters. Compared to WayFast~\cite{Gasparino2022wayfast} that uses the expected traction values (a risk-neutral special case of our second method), our approach can produce behaviors that are more risk-averse by adjusting the worst-case quantiles used to compute the CVaR of traction.

\subsection{Uncertainty Estimation for Neural Networks}
Data collection in practice is often expensive and limited in diversity, so it is important to know when a learned model cannot be trusted (e.g., due to input features that are significantly different from training data).
NN uncertainty estimation is well studied in the machine learning literature (e.g., see survey~\cite{gawlikowski2021survey}), where representative techniques include Monte Carlo dropouts~\cite{gal2016dropout}, ensembles~\cite{osband2016deep}, and single-pass methods~\cite{natpn}. Most similar to the single-pass technique~\cite{natpn} that only requires a single neural network evaluation to estimate uncertainty, our work tries to capture the aleatoric uncertainty (the inherent process uncertainty that cannot be reduced with more data) by predicting a distribution, and the epistemic uncertainty (the model uncertainty that can be reduced with more data) by leveraging the latent space density of NN. Compared to using dropouts or ensembles, single-pass methods are attractive, because they do not require taking extra samples or using more memory. 

\section{Problem Formulation}
We consider the problem of motion planning for a wheeled vehicle whose dynamics depend on the underlying terrains, where the traction values are uncertain due to imperfect sensing.
Therefore, we model traction values as random variables whose distributions can be learned empirically.

\subsection{Unicycle Model with Uncertain Traction Parameters}
Consider the discrete time system:
\begin{equation}\label{eq:dynamics}
    \mathbf{x}_{t+1} = F(\mathbf{x}_t, \mathbf{u}_t, \Param_{t}),
\end{equation}
where $\mathbf{x}_t\in\setX\subseteq\R^n$ is the state vector, $\mathbf{u}_t\in\R^m$ is the control input, and $\Param_{t}\in\setParam\subseteq \R^q$ is the parameter vector that captures the terrain traction. 
For each mission, terrain-dependent parameter vectors are sampled from the ground truth distribution $\Param \sim p^*(\cdot \vert \mathbf{o}_{\mathbf{x}})$ for every $\mathbf{x}\in\setX$ and the associated terrain features $\mathbf{o}_{\mathbf{x}}\in\setO$.
For concreteness, we use the following unicycle model
\begin{equation}\label{eq:unicycle}
    \begin{bmatrix}
        p_{t+1}^{x}\\
        p_{t+1}^{y}\\
        \theta_{t+1}
    \end{bmatrix} = 
    \begin{bmatrix}
        p_{t}^{x}\\
        p_{t}^{y}\\
        \theta_{t}
    \end{bmatrix} + \Delta \cdot
    \begin{bmatrix}
        \psi_1 \cdot v_t \cdot \cos{(\theta_t)}\\
        \psi_1 \cdot v_t \cdot \sin{(\theta_t)}\\
        \psi_2 \cdot \omega_t
    \end{bmatrix},
\end{equation}
where $\mathbf{x}_t=[p_{t}^{x}, p_{t}^{y}, \theta_{t}]\tr$ contains the X, Y positions and yaw, $\mathbf{u}_t=[v_t, \omega_t]\tr$ contains the linear and angular velocities, $\Param_{t}=[\psi_1, \psi_2]\tr$ contains the linear and angular traction values $0\leq \psi_1, \psi_2 \leq1$, and $\Delta>0$ is the time interval. 
Intuitively, traction captures how much of the commanded velocities can be achieved and is a good indicator for terrain traversability for fast off-road navigation.

\subsection{Motion Planning}
As this work focuses on achieving fast navigation to a given goal position, we adopt and modify the minimum-time formulation used in~\cite{cai2022risk}, but any other task-specific objectives can be used instead. Given initial state $\mathbf{x}_0$ and goal position $\mathbf{p}^{\text{goal}}\in\R^2$, the problem of finding a sequence of control $\mathbf{u}_{0:T-1}$ can be written as
\begin{align}
\min_{\mathbf{u}_{0:T-1}} \quad & C(\mathbf{x}_{0:T}) \defeq \phi(\mathbf{x}_T)+\sum_{t=0}^{T-1} q(\mathbf{x}_t),\label{eq:nominal_obj}\\
\textrm{s.t.} \quad & \mathbf{x}_{t+1} = F(\mathbf{x}_t, \mathbf{u}_t, \Param_{t}),\quad\forall t\in\{0,\dots,T-1\},
\end{align}
where $\phi(\mathbf{x}_T)$ and $q(\mathbf{x}_t)$ are the terminal cost and the stage cost:
\begin{align}
\phi(\mathbf{x}_T) &= \frac{\norm{\mathbf{p}^{\text{goal}} - \mathbf{p}_T }}{s^{\text{default}}}\left( 1-\done{\mathbf{x}_{0:T}} \right), \\
q(\mathbf{x}_t) &= \Delta \left( 1-\done{\mathbf{x}_{0:t}} \right) + w^{\text{dist}}\cdot \norm{\mathbf{p}^{\text{goal}} - \mathbf{p}_t },
\end{align}
where $s^{\text{default}}$ is the default speed for estimating time-to-go at the end of the rollout, $\Delta$ is the sampling duration, $\mathbf{p}_t$ is the robot position at $t$, and $w^{\text{dist}}>0$ is the weight for penalizing distance from the goal. To avoid accumulating costs after robot reaches the goal, we use an indicator function $\done{\mathbf{x}_{0:t}}$ that returns $1$ when any state $\mathbf{x}_{\tau}$ has reached $\mathbf{p}^{\text{goal}}$ for $0\leq\tau\leq T$, and returns $0$ otherwise. Intuitively, the objective encourages the robot to arrive at the goal location as quickly as possible.

While the problem~\eqref{eq:nominal_obj} can be optimized via non-linear optimization techniques such as Model Predictive Path Integral control (MPPI~\cite[Algorithm~2]{williams2017information}), the main challenge is that the \textit{terrain traction $\Param$ is uncertain}. To address this issue, existing techniques try to learn the expected traction, or use nominal traction while penalizing undesirable terrains manually. However, these approaches are either risk-neutral or require human expertise in cost design.
In this work, we propose to learn the full traction distribution empirically (Sec.~\ref{sec:traversability}) that can be used to design risk-aware costs with easy-to-tune risk tolerance (Sec.~\ref{sec:planning}).

\section{Traversability Model}\label{sec:traversability}
\begin{figure*}[t]
	\centering
	\includegraphics[width=\linewidth, trim={0cm 0cm 0cm 0cm},clip]{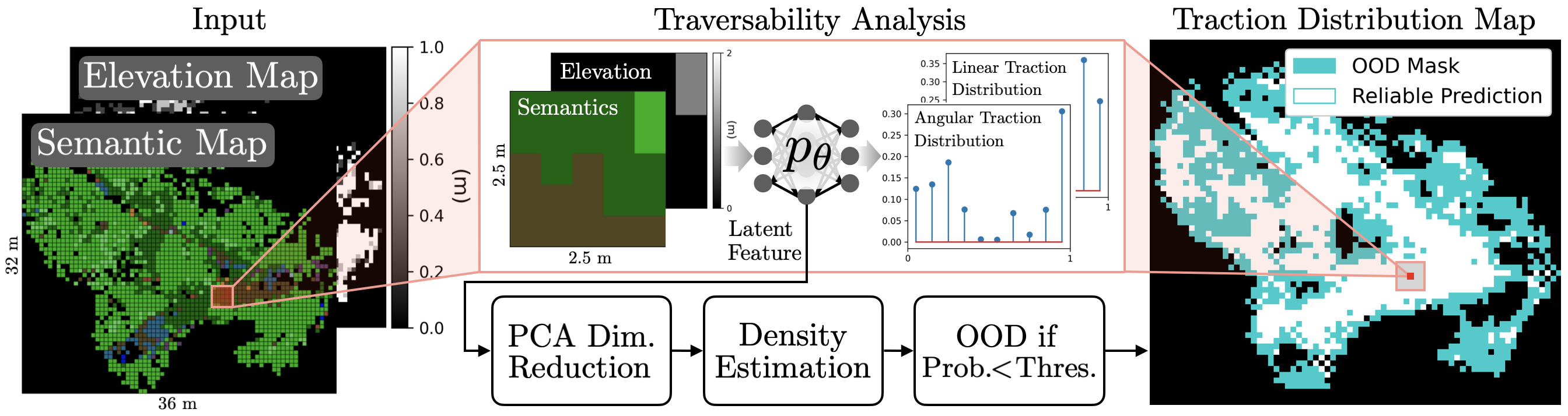}
	\caption{Overview of the proposed traversability analysis procedure that shows the real-world terrain features and the actual empirical traction distribution learned by the trained NN.
	The proposed method captures both the aleatoric uncertainty via the predicted distribution, and the epistemic uncertainty via the latent space density. In a sliding-window fashion, the trained NN takes in local semantic and elevation features to predict linear and angular traction distributions. By learning categorical distributions over the discretized traction values, the NN is able to capture rich terrain properties that are not necessarily Gaussian. To identify terrain features that are significantly different from the training data, we fit a density estimator in the trained NN's latent space, such as a Gaussian Mixture Model (GMM), in order to obtain the likelihood of any input terrain during deployment. If the likelihood of input terrain features is below a certain threshold, the terrain is deemed out-of-distribution (OOD) and later avoided during planning via auxiliary penalties.
	}
	\label{fig:traversability_analysis_pipeline} 
	\vspace*{-0.15in}
\end{figure*}

In this section, we introduce how to learn traction distribution (aleatoric uncertainty) and how to leverage density estimation in the trained NN's latent space for detecting unfamiliar terrains (epistemic uncertainty). An overview of the entire traversability analysis procedure is shown in Fig.~\ref{fig:traversability_analysis_pipeline}.

\begin{figure}[t]
	\centering
	\includegraphics[width=\linewidth, trim={0cm 0cm 0cm 0cm},clip]{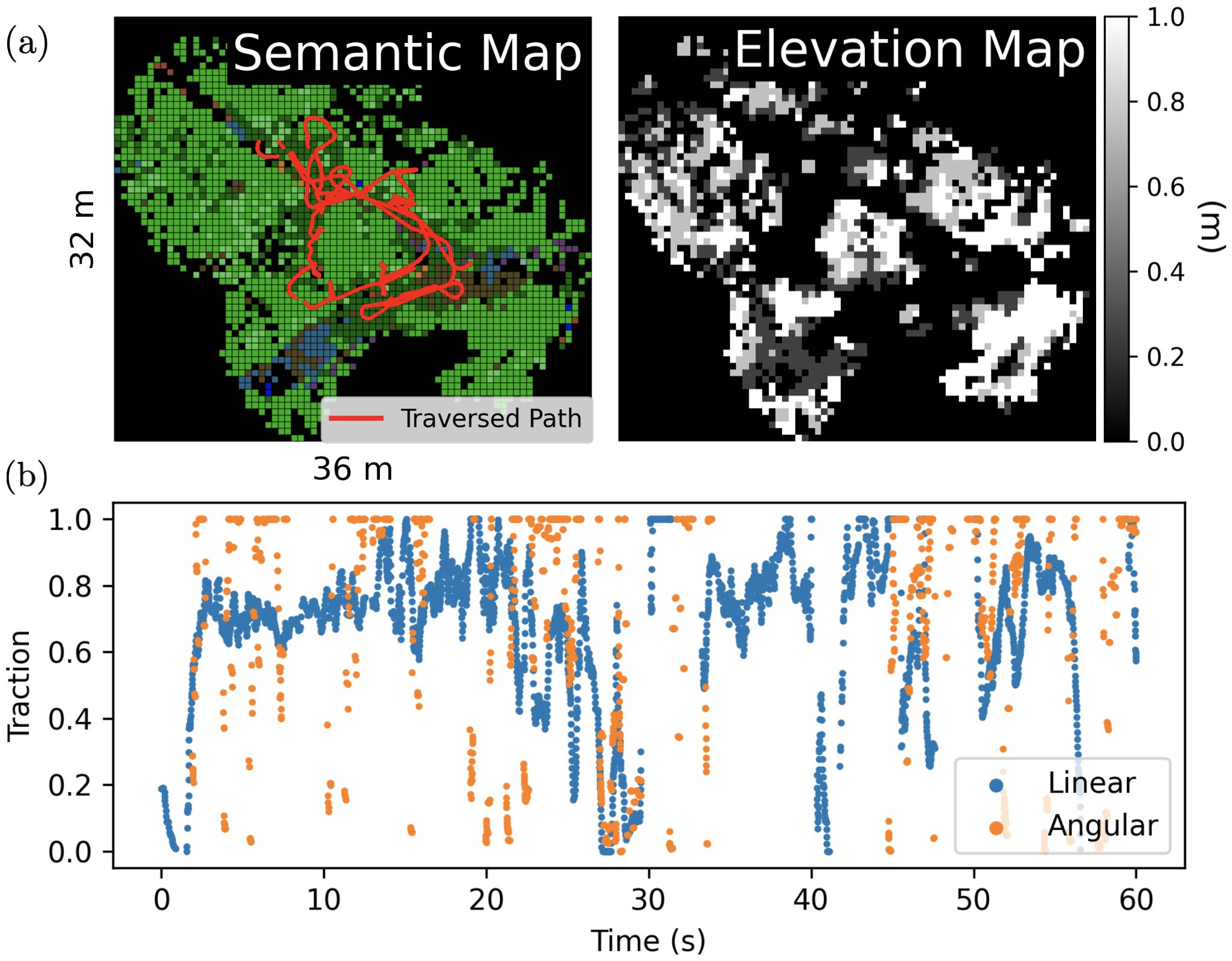}
	\caption{Illustration of a real world data collection used to obtain terrain features and traction values.
	(a) The robot was manually driven for 10 minutes over different terrains while building a semantic octomap from which a projected 2D semantic map and an elevation map can be extracted. Example terrain types include vegetation (light green), grass (dark green) and mulch (brown).
	(b) Only a subset of the collected linear and angular traction values are shown for clarity. Note that discontinuity in traction values occurred when linear or angular commands were not sent.
	}
	\label{fig:data_collection}
\vspace*{-0.15in}
\end{figure}

\subsection{Terrain-Dependent Traction Distribution}

Given the set of system parameters $\setParam$ and the set of terrain features $\setO$, we want to model the conditional distribution
\begin{equation}
p_{\btheta}(\Param \mid \mathbf{o}):\setParam \mid  \setO \rightarrow \R, \label{eq:cond_distribution}
\end{equation}
where $p_{\btheta}$ is a probability distribution parameterized by $\btheta$, which in practice can be learned by a NN using empirically collected dataset $\{(\mathbf{o}, \Param)_k\}_{k=1}^{K}$ where $K>0$. A real world example can be found in Fig.~\ref{fig:data_collection} where a Clearpath Husky was manually driven in a forest to build an environment model and collect traction data. Note that the semantic and geometric information about the environment can be built by using a semantic octomap~\cite{asgharivaskasi2021active} that temporally fuses semantic point clouds.
We used PointRend~\cite{kirillov2020pointrend} trained on RUGD off-road navigation dataset~\cite{RUGD2019IROS} to segment RGB images and subsequently projected the semantics onto lidar point clouds. 
To estimate the true linear and angular velocities of the robot, we could not rely on the wheel-encoders due to wheel slips. Therefore, we used direct lidar odometry~\cite{chen2022direct} that produced accurate pose estimates even when driving through tall grass, and the resultant pose estimates were fused with IMU measurements in an extended Kalman filter to obtain the high-rate velocity estimates. The velocity estimates were further filtered to reduce noise due to bumpy terrains. Finally, the traction values were computed as the ratios between the estimated and the commanded velocities and stored for offline training.

Given terrain features, traversed path and estimated traction values, we can train a NN (e.g., with convolutional layers followed by fully connected layers) to map local terrain features to categorical distributions over discretized traction values in order to capture rich terrain properties. %
To facilitate planning, we store learned traction distributions in a map $\setM_{\btheta}$, where each cell $\setM_{\btheta}^{h,w}$ indexed by row $h$ and height $w$ stores the traction distribution $p_{\btheta}^{h,w}:\setParam \rightarrow \R$ that has already been conditioned on the associated terrain features.

\subsection{Density-Based Detector for Unfamiliar Terrain Features}\label{sec:ood_detector}

As the learned traction predictor is trained on limited data, its predictions based on terrain features significantly different from training data are unreliable, which can lead to degraded navigation performance. Therefore, we fit a density estimator in the latent space of the trained NN in order to use the predicted likelihood as a measure of model uncertainty.

We first apply principal component analysis (PCA) to the latent space features that correspond to all the terrain features observed during training $\boldsymbol{O}^{\text{train}}=\{\mathbf{o}_k\}_{k=1}^{K}$. Next, we fit a Gaussian Mixture Model (GMM) for the entire training dataset in the reduced latent space, where the likelihood of observing a particular feature $\mathbf{o}_k$ is denoted as $p_{\btheta}^{\text{latent}}(\mathbf{o}_k)$. We design a simple confidence score $g$ based on the log-likelihood of the query data normalized between the maximum and minimum log-likelihood observed in training data:
\begin{align}
    g(\mathbf{o}) &= \frac{p_{\btheta}^{\text{latent}}(\mathbf{o}) - p^{\text{min}}}{p^{\text{max}}-p^{\text{min}}}, \label{eq:gmm_conf}\\
    p^{\text{max}}&= \max_{\mathbf{o}'\in\boldsymbol{O}^{\text{train}}} p_{\btheta}^{\text{latent}}(\mathbf{o}'),\\
    p^{\text{min}}&= \min_{\mathbf{o}'\in\boldsymbol{O}^{\text{train}}} p_{\btheta}^{\text{latent}}(\mathbf{o}').
\end{align}
Note that $g(\mathbf{o})$ is not limited to $[0,1]$ and lower values indicate that the terrain features are less similar to the training data. With the NN trained in the environment shown in Fig.~\ref{fig:data_collection}, we project the latent space features to the first 2 principle components and fit a GMM with 2 clusters. During deployment, terrain features with confidence score below some threshold $g^{\text{thres}}\in[0,1]$ are deemed out-of-distribution (OOD) and the OOD terrains should be explicitly avoided during planning via auxiliary penalties.
This strategy improves navigation success rate when the NN is deployed in an environment unseen during training (see Sec.~\ref{sec:exp_bag_sim}).

\section{Planning with Learned Traction Distribution}\label{sec:planning}
Given the learned traction model, we propose two risk-aware cost formulations in order to generate control input that is less likely to lead to worst-case failures. Note that the proposed cost formulations can substitute the nominal objective in~\eqref{eq:nominal_obj} and the problem can still be solved normally using nonlinear optimization technique such as MPPI.

\subsection{Conditional Value at Risk (CVaR)}

We adopt the Conditional Value at Risk (CVaR) as a risk metric because it satisfies a group of axioms important for rational risk assessment~\cite{majumdar2020should}, but the conventional definition assumes the worst-case occurs at the right tail of the distribution. In this work, we define CVaR at both the right and left tails (see Fig.~\ref{fig:cvar_definition}) at level $\alpha\in(0,1]$ for the random variable $Z$ and its possible realization $z\in\R$ as follows:
\begin{align}
    \rcvar{\alpha}{Z} &\defeq \frac{1}{\alpha} \int_{0}^{\alpha} \rvar{\tau}{Z}\ d\tau, \\
    \lcvar{\alpha}{Z} &\defeq \frac{1}{\alpha} \int_{0}^{\alpha} \lvar{\tau}{Z}\ d\tau,
\end{align}
where the right and left Values at Risk are defined as:
\begin{align}
\rvar{\alpha}{Z} &\defeq \min \{ z \mid p(Z > z) \leq \alpha\}, \\
\lvar{\alpha}{Z} &\defeq \max \{ z \mid p(Z < z) \leq \alpha\}.
\end{align}
Intuitively, $\rcvar{\alpha}{Z}$ and $\lcvar{\alpha}{Z}$ capture the expected outcomes that fall in the right tail and left tail of the distribution, respectively, where each tail occupies $\alpha$ portion of the total probability. Notice that either definition of CVaR produces the mean when $\alpha=1$.

\subsection{Risk-Aware Cost Formulations}\label{sec:risk_aware_cost}
\begin{figure}[t]
	\centering
	\includegraphics[width=0.7\linewidth, trim={0cm 0cm 0cm 0cm},clip]{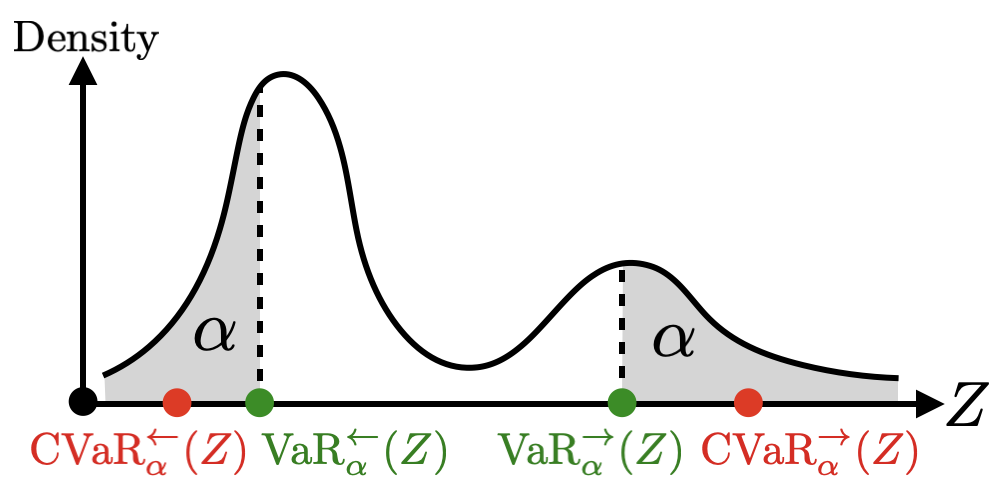}
	\caption{This work defines two versions of Conditional Value at Risk (CVaR) to capture the worst-case expected values at either the left tail as $\lcvar{\alpha}{Z}$ or the right tail as $\rcvar{\alpha}{Z}$ for some random variable $Z$, where the worst-case scenarios constitute $\alpha\in(0,1]$ portion of total probability. The left-tail and right-tail Values at Risk (VaR) are defined as $\lvar{\alpha}{Z}$ and $\rvar{\alpha}{Z}$. Note that the right-tail definitions are suitable for costs to be minimized, and the left-tail definitions are suitable for low traction values.}
	\label{fig:cvar_definition}
\vspace*{0.10in}
	\centering
	\includegraphics[width=\linewidth, trim={0cm 0cm 0cm 0cm},clip]{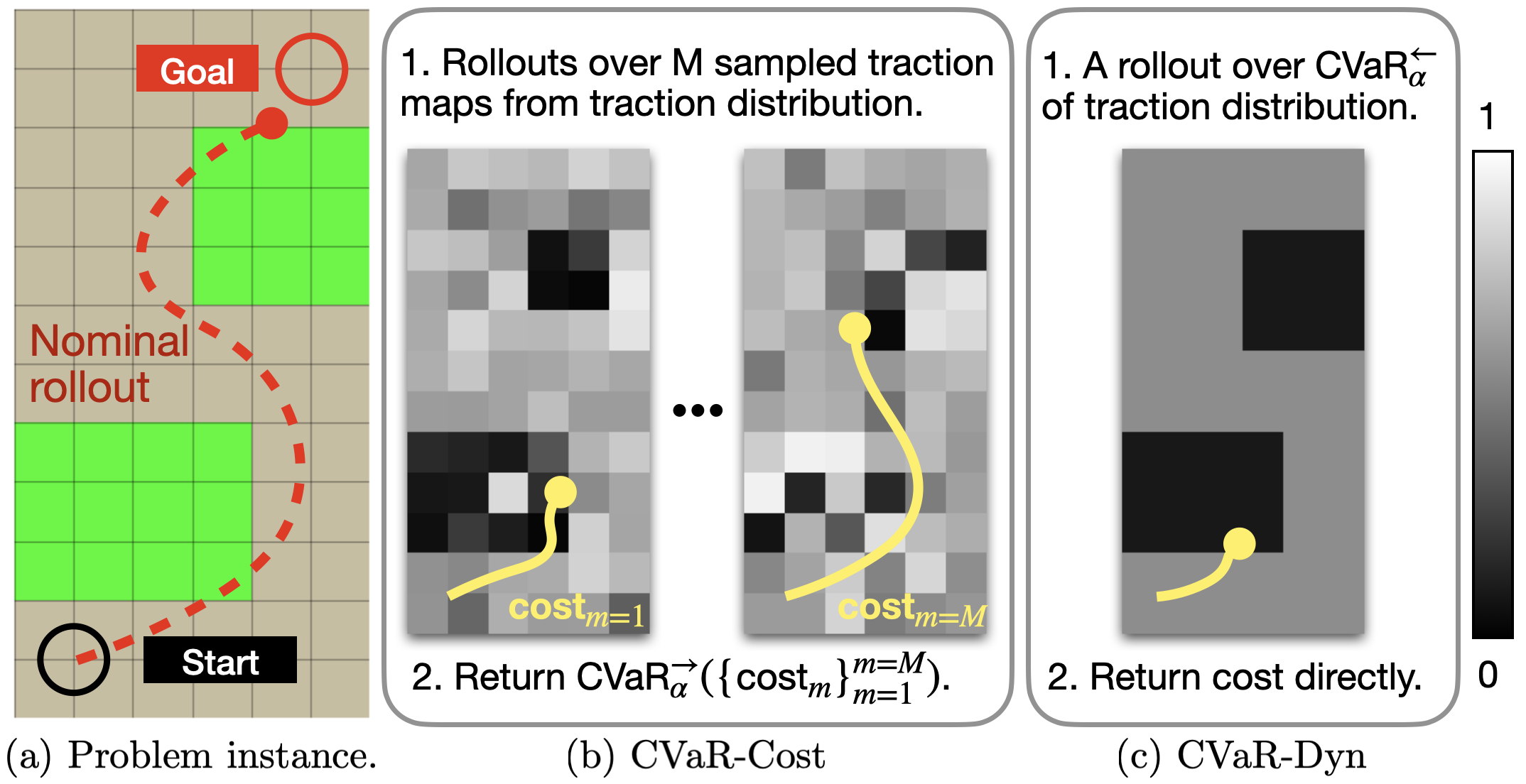}
	\caption{
	    Illustrations of the proposed risk-aware costs in a toy problem with known traction distributions for vegetation and dirt terrains as shown in Fig.~\ref{fig:sim_benchmark_setup}. Linear and angular traction models are assumed equal.  
	    (a) The state rollout based on nominal dynamics (i.e., no slip) does not account for uncertain traction. 
	    (b) \cvarcost{} computes the expected costs in the right $\alpha$-quantile by evaluating the given control sequence over $M$ traction map samples. 
	    (c) \cvardyn{} requires only a single rollout over the traction map that contains the expected traction in the left $\alpha$-quantile.}
	\label{fig:cost_computation}
 \vspace*{-0.15in}
\end{figure}

In order to account for the risk of obtaining high cost due to the uncertain system parameters $\Param_t$, we propose to optimize two modified versions of the cost function~\eqref{eq:nominal_obj}. Fig.~\ref{fig:cost_computation} gives a high-level illustration of the core ideas behind the two cost functions.

\subsubsection{Worst-Case Expected Cost (\cvarcost{})}

Given the control sequence $\mathbf{u}_{0:T-1}$, we want to evaluate the worst-case expected value for the nominal objective $C$~\eqref{eq:nominal_obj} due to uncertain terrain traction. First, we sample from the traction distribution map $\setM_{\btheta}$ to obtain $M>0$ traction  maps that contain samples of traction values from distribution $p_{\btheta}^{h,w}$ in every map cell indexed by height $h$ and width $w$ for each sample index $m\in\{1,\dots,M\}$. Next, we compute the empirical right-tail CVaR of the performances of the $M$ rollouts:
\begin{equation}
    c^{\text{\cvarcost{}}} \defeq \rcvar{\alpha}{\{C(\mathbf{x}^m_{0:T})\}_{m=1}^M}, \label{eq:cvar_cost}
\end{equation}
where the $m$-th state rollout $\mathbf{x}^m_{0:T}$ follows
\begin{equation}
 \mathbf{x}_{t+1}^m  = F(\mathbf{x}_t^m, \mathbf{u}_t, \Param_t^{m}),\quad \mathbf{x}_{0}^m = \mathbf{x}_{0},
\end{equation}
for $t\in\{0,\dots,T-1\}$. The traction parameter $\Param_t^m$ is queried in the $m$-th sampled traction map at state $\mathbf{x}_t^m$. Note that this approach is inspired by~\cite{wang2021adaptive}, but we additionally handle terrain-dependent distributions of parameters and the sampling of traction maps. For better real-time performance, the sampled traction maps can be reused for evaluating different control sequences.

\subsubsection{Worst-Case Expected System Parameters (\cvardyn{})}
The procedure for evaluating a control sequence over a large number of sampled traction maps can be efficiently parallelized on GPUs, but the computational overhead can still grow prohibitively when considering many control sequences. Therefore, we propose an alternative cost design that accounts for the worst-case expectation of the traction values in the dynamical model.

Given the control sequence $\mathbf{u}_{0:T-1}$, we evaluate the nominal mission objective $C$~\eqref{eq:nominal_obj} based on the state rollout simulated with the worst-case expected traction, i.e., 
\begin{equation}
    c^{\text{\cvardyn{}}} \defeq C(\overline{\mathbf{x}}_{0:T}), \label{eq:cvar_dyn}
\end{equation}
where the state rollout follows
\begin{equation}
 \overline{\mathbf{x}}_{t+1}  = F(\overline{\mathbf{x}}_t, \mathbf{u}_t, \overline{\Param}_t),\quad \overline{\mathbf{x}}_{0} = \mathbf{x}_{0},
\end{equation}
for $t\in\{0,\dots,T-1\}$ and the worst-case expected traction $\overline{\Param}_{t}$ is computed based on the corresponding traction distribution at some row $h$ and height $w$ determined by state $\overline{\mathbf{x}}_t$:
\begin{equation}
    \overline{\Param}_{t}=
    \begin{bmatrix}
    \lcvar{\alpha}{\Psi_1} \\
    \lcvar{\alpha}{\Psi_2}
    \end{bmatrix},\ 
    \begin{bmatrix}
    \Psi_1 \\
    \Psi_2
    \end{bmatrix}\sim p_{\btheta}^{h,w}.
\end{equation}
When $\alpha=1$, the expected values of the traction parameters are used, equivalent to the approach in~\cite{Gasparino2022wayfast}. However, as the results in Sec.~\ref{sec:results:semantic_map} show for a go-to-goal task, planning with the worst-case expected traction can improve navigation performance when the traction distribution is not Gaussian.

\begin{remark}
The proposed costs~\eqref{eq:cvar_cost} and~\eqref{eq:cvar_dyn} provide intuitive notions of risk that depend on the worst-case expected cost and terrain traction. Moreover, they are simple to tune with a single risk parameter $\alpha$, avoiding the need to manually design weights for a potentially large variety of terrains. 
\end{remark}

\begin{remark}
While adding auxiliary penalties for trajectories entering low-traction terrains can generate similar risk-aware behaviors achievable by our approach, we show that using the proposed costs~\eqref{eq:cvar_cost} and~\eqref{eq:cvar_dyn} leads to solutions with better trade-offs between success rate and time-to-goal than ones achieved by using the nominal objective augmented with terrain penalties (see Fig.~\ref{fig:veg_penalty}). However, the highest success rate achieved by using the proposed cost formulations are lower. 
Combining the best of both worlds, using the proposed costs with auxiliary penalties for undesirable terrains, such as OOD terrains, lead to better performance (see Sec.~\ref{sec:exp_bag_sim}).
\end{remark}

\section{Simulation Results}
Using simulated semantic environments, we show that the proposed methods outperform existing approaches~\cite{Gasparino2022wayfast, cai2022risk} that either assume no slip or use expected traction. Moreover, we discuss the advantages and limitations of our proposed risk-aware costs compared to auxiliary penalties for low-traction terrains. 
Lastly via simulations based on real-world traction data, we show that avoiding OOD terrains improves navigation success rate, and using our approach with auxiliary cost for OOD terrains improves time-to-goals.
To prevent simulations from running indefinitely when the robot encounters near-zero traction, we impose time limits (selected based on mission difficulties) that are much longer than the average time required to complete the missions.

\begin{figure*}[t]
	\centering
	\includegraphics[width=0.92\linewidth, trim={0cm 0cm 0cm 0cm},clip]{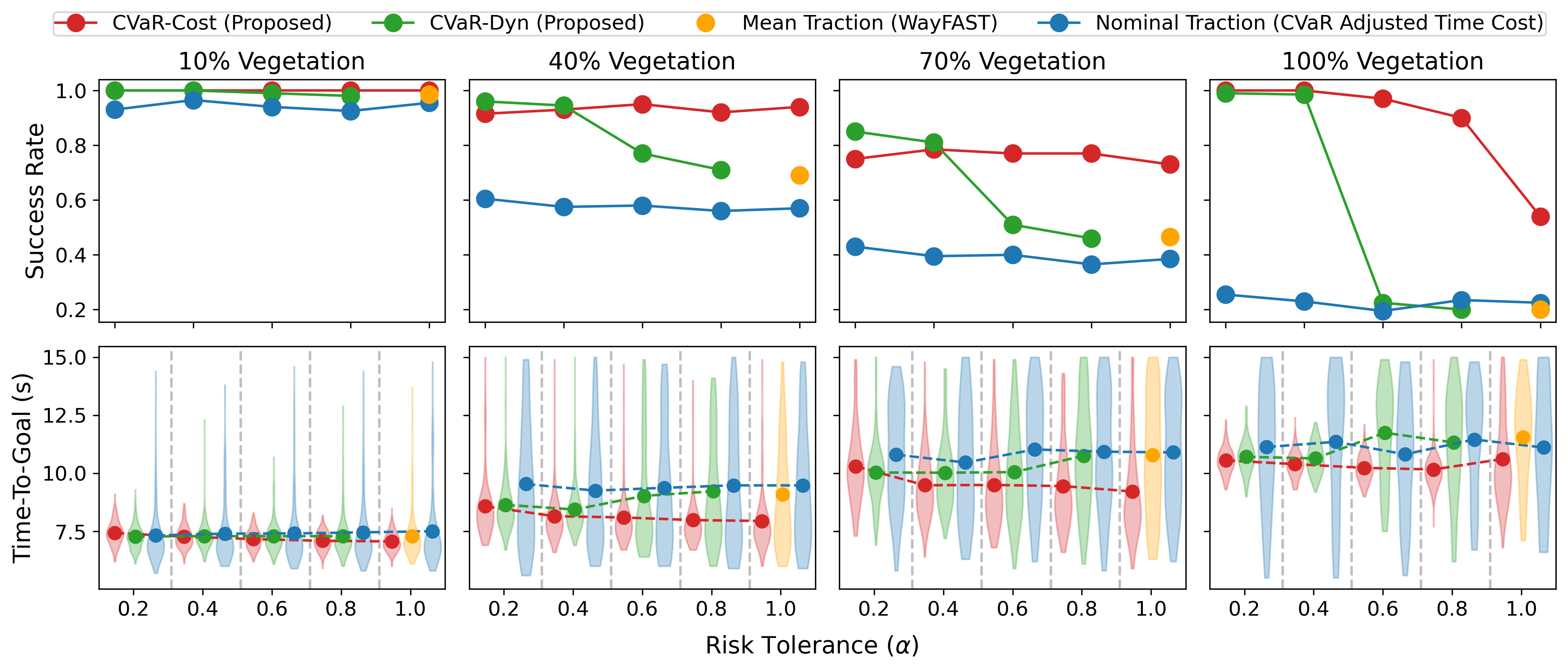}
	\caption{Results comparing the proposed methods (\cvarcost{} and \cvardyn{} in Sec.~\ref{sec:risk_aware_cost}) against existing methods based on the expected traction (WayFAST~\cite{Gasparino2022wayfast}) and the method that assumes nominal traction~\cite{cai2022risk} (i.e., no slip). Note that a mission is successful if the goal is reached within 15~s.
    Overall, the \cvarcost{} planner always maintains better success rate and time-to-goal than WayFAST and the method using nominal traction. As the risk tolerance increases, the \cvarcost{} planner becomes more optimistic and achieves lower time-to-goal.
    When the risk tolerance is sufficiently low (e.g., $\alpha=0.2$), the second proposed planner \cvardyn{} achieves similar or better success rate and time-to-goal compared to the \cvarcost{} planner. %
    }
	\label{fig:sim_benchmark} \vspace*{-0.15in}
\end{figure*}

\subsection{Implementation Details}
In order to solve the optimization problem~\eqref{eq:nominal_obj}, we adopt MPPI~\cite[Algorithm~2]{williams2017information} to generate control for achieving fast navigation to goal. This approach is attractive because it is derivative-free, parallelizable on GPU, and works with our proposed risk-aware cost formulations~\eqref{eq:cvar_cost} and~\eqref{eq:cvar_dyn}. The MPPI planners run in a receding horizon fashion with 100 time steps and each step is 0.1~s. The max linear and angular speeds are capped at 3~m/s and $\pi$~rad/s, and the noise standard deviations for the control signals are 2~m/s and 2~rad/s. The number of control rollouts is 1024 and the number of sampled traction maps is 1024 (only applicable for the \cvarcost{}). We use probability mass functions with 20 uniform bins to approximate the parameter distribution.
The \cvarcost{} planner is the most expensive to compute, but it is able to re-plan at 15~Hz while sampling new control actions and maps with dimension of $200\times 200$. Planners that do not sample traction maps can be executed at over 50~Hz.
A computer with Intel Core i9 CPU and Nvidia GeForce RTX 3070 GPU is used for the simulations, where majority of the computation happens on the GPU.

\subsection{Simulated Semantic Environments}\label{sec:results:semantic_map}

\begin{figure}[t]
     \centering
    \includegraphics[width=\linewidth, trim={0cm 0cm 0cm 0cm},clip]{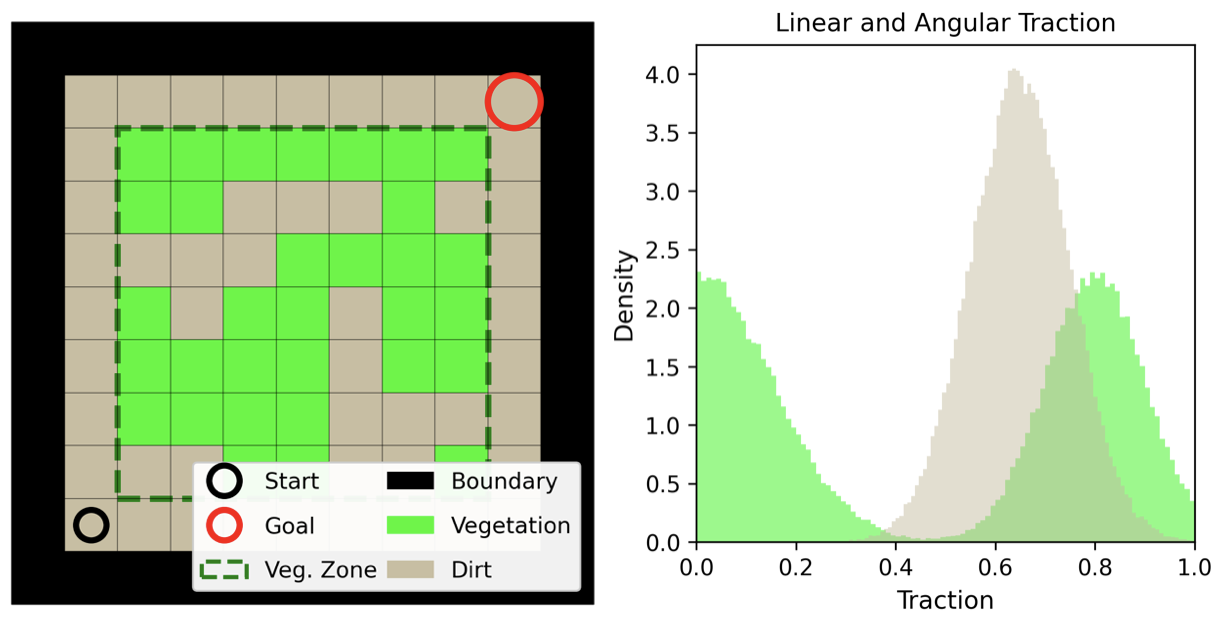}
        \caption{The simulation environment where a robot has to move from start to goal as fast as possible within the bounded arena. Linear and angular traction parameters share the same distribution for simplicity. Vegetation terrain patches are randomly sampled at the center.}
        \label{fig:sim_benchmark_setup}
  \vspace*{-0.15in}
\end{figure}

We consider a grid world scenario where ``dirt'' and ``vegetation'' cells have known traction distributions, as shown in Fig.~\ref{fig:sim_benchmark_setup}. Vegetation patches are randomly spawned with increasing probabilities at the center of the arena, and a robot may experience significant slow-down for certain vegetation cells due to vegetation's bi-modal distribution. The mission is deemed successful if the goal is reached within 15~s.

Overall, we sample $40$ different semantic maps and $5$ random realizations of traction parameters for every semantic map. The traction parameters are drawn before starting each trial and remain fixed. The benchmark results can be found in Fig.~\ref{fig:sim_benchmark}, where we compare the two proposed costs with existing methods, namely WayFAST~\cite{Gasparino2022wayfast} that uses the expected traction and the technique in~\cite{cai2022risk} that assumes nominal dynamics while adjusting the time cost with the CVaR of linear traction. The takeaway is that the proposed methods outperform the two existing ones by accounting for the worst-case expected cost and traction.
Notably, although the \cvardyn{} planner does not sample from the entire parameter distribution, it achieves similar performance to the \cvarcost{} planner when $\alpha$ is set sufficiently low. The poor performance of the \cvarcost{} planner can be attributed to the difficulty of estimating CVaR from samples in general when $\alpha$ is low. However, the \cvardyn{} planner makes conservative assumption about the dynamics, so it does not out-perform \cvarcost{} in achieving low time-to-goal.

\begin{figure}[t]
     \centering
    \includegraphics[width=0.95\linewidth, trim={0cm 0cm 0cm 0cm},clip]{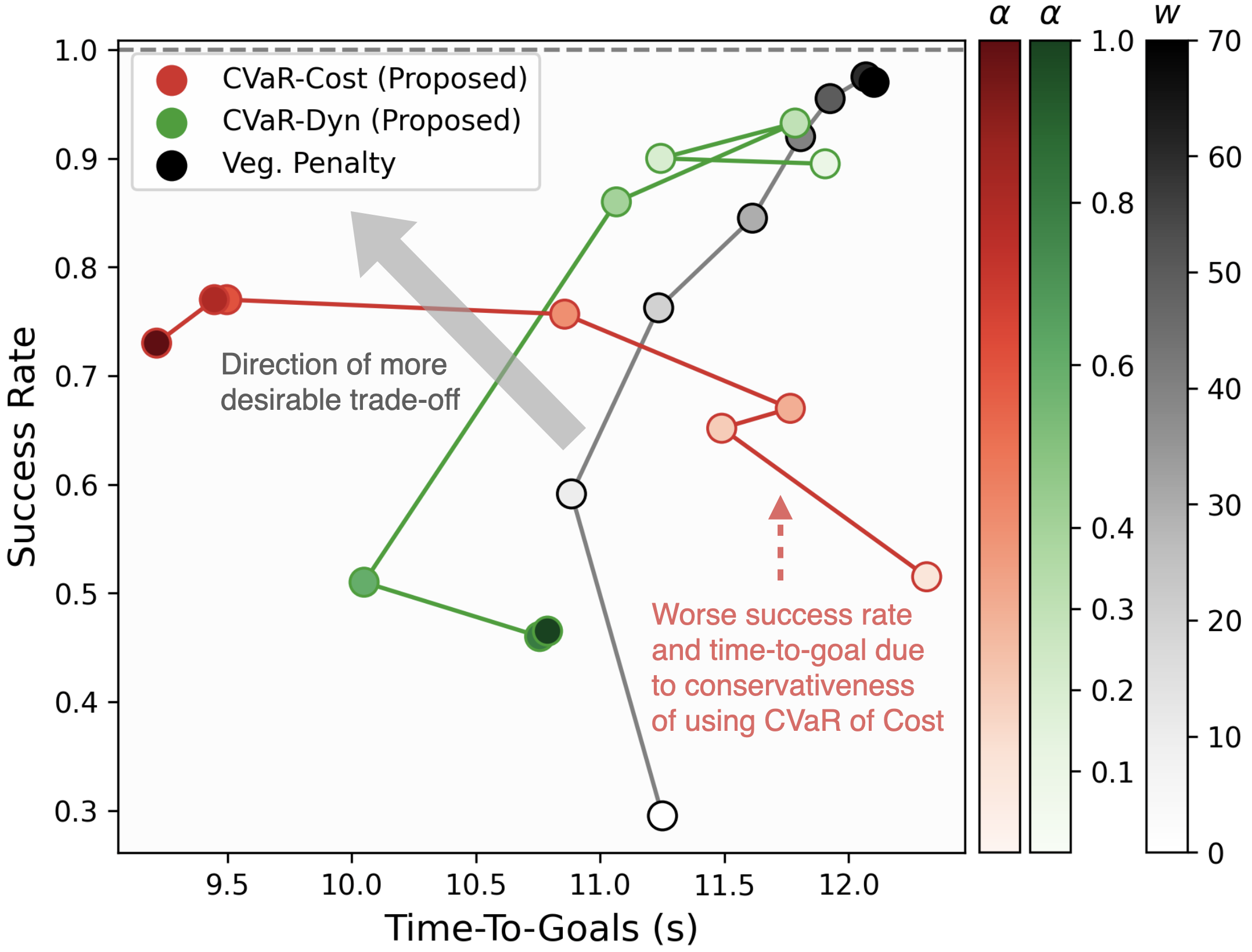}
        \caption{
        In the most challenging scenario of $70\%$ vegetation, solutions obtained by \cvardyn{} (green) achieve a better trade-off between success rate and time-to-goal than ones achieved by adding auxiliary penalty $w>0$ for states entering vegetation terrains (black) by being in the upper left of the figure. However, the success rate of \cvardyn{} plateaus as $\alpha$ decreases due to shorter state rollouts that lead to local minima.
        Although the \cvarcost{} planner (red) achieves the best time-to-goal with high $\alpha$, its performance suffers significantly from the conservativeness of using CVaR of the objective, which leads to worsening success rate and time-to-goal as $\alpha$ lowers.
        The conservativeness of the proposed costs can be mitigated by adding auxiliary penalties for undesirable terrains to improve navigation performance when domain knowledge is available (see Sec.~\ref{sec:exp_bag_sim}).
        }
        \label{fig:veg_penalty}
  \vspace*{-0.15in}
\end{figure}

To compare our cost formulations against the nominal objective with auxiliary stage costs that penalize states in vegetation cells, we focus on the most challenging setting with 70\% vegetation where it is easy to get stuck in local minima. The benchmark result is shown in Fig.~\ref{fig:veg_penalty} where we compare the trade-offs between success rate and time-to-goal achieved by different methods. Although there exists risk tolerance $\alpha$ that allows \cvarcost{} and \cvardyn{} to obtain better trade-off than having vegetation penalties, their conservativeness in considering the CVaR of cost and traction prevents them from achieving solutions with higher success rate. 
Therefore, when domain knowledge is available, adding auxiliary costs can help achieve much higher success rate desirable in practice, but tuning the cost may be challenging when a large variety of terrains exist.

\subsection{Simulated Traction Based on Real-World Data}
\label{sec:exp_bag_sim}

\begin{figure}[t!]
     \centering
    \includegraphics[width=0.9\linewidth, trim={0cm 0cm 0.cm 0cm},clip]{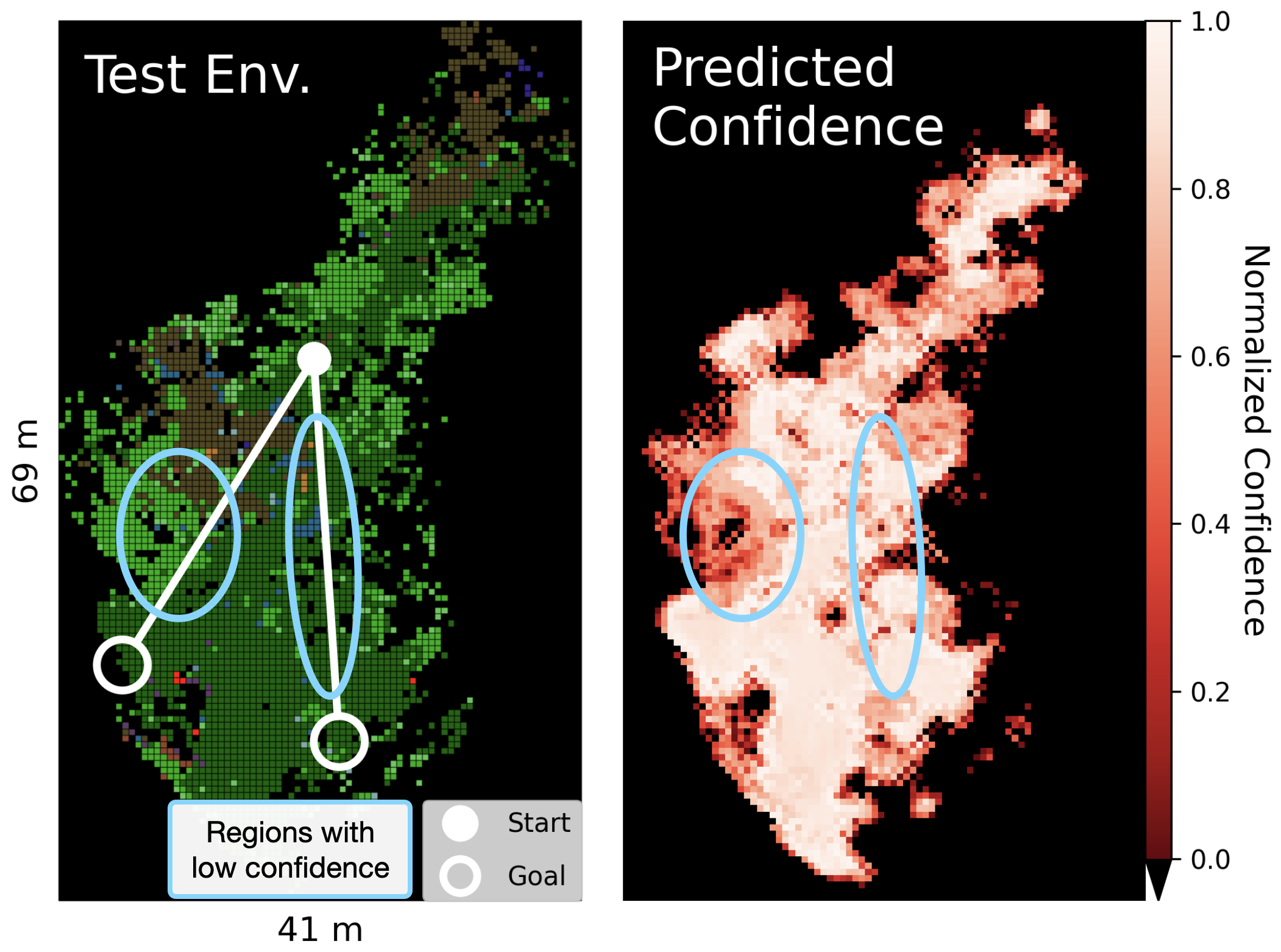}
        \caption{In a test environment unseen during training (left), the robot has to reach two goals selected to highlight the danger of using unreliable NN predictions due to high model uncertainty. (Right) The GMM-based confidence score~\eqref{eq:gmm_conf} indicates the amount of model uncertainty for the predicted traction distribution in each map location, where unknown terrains and known terrains with negative scores are shown in black. Note that the brown  semantic region (mulch) at the top has confidence below zero due to the presence of unknown cells, in contrast to the brown semantic region to the left with much fewer unknown cells.}
        \label{fig:test_env}
\vspace*{0.10in}
	\centering
	\includegraphics[width=\linewidth, trim={0cm 0cm 0cm 0cm},clip]{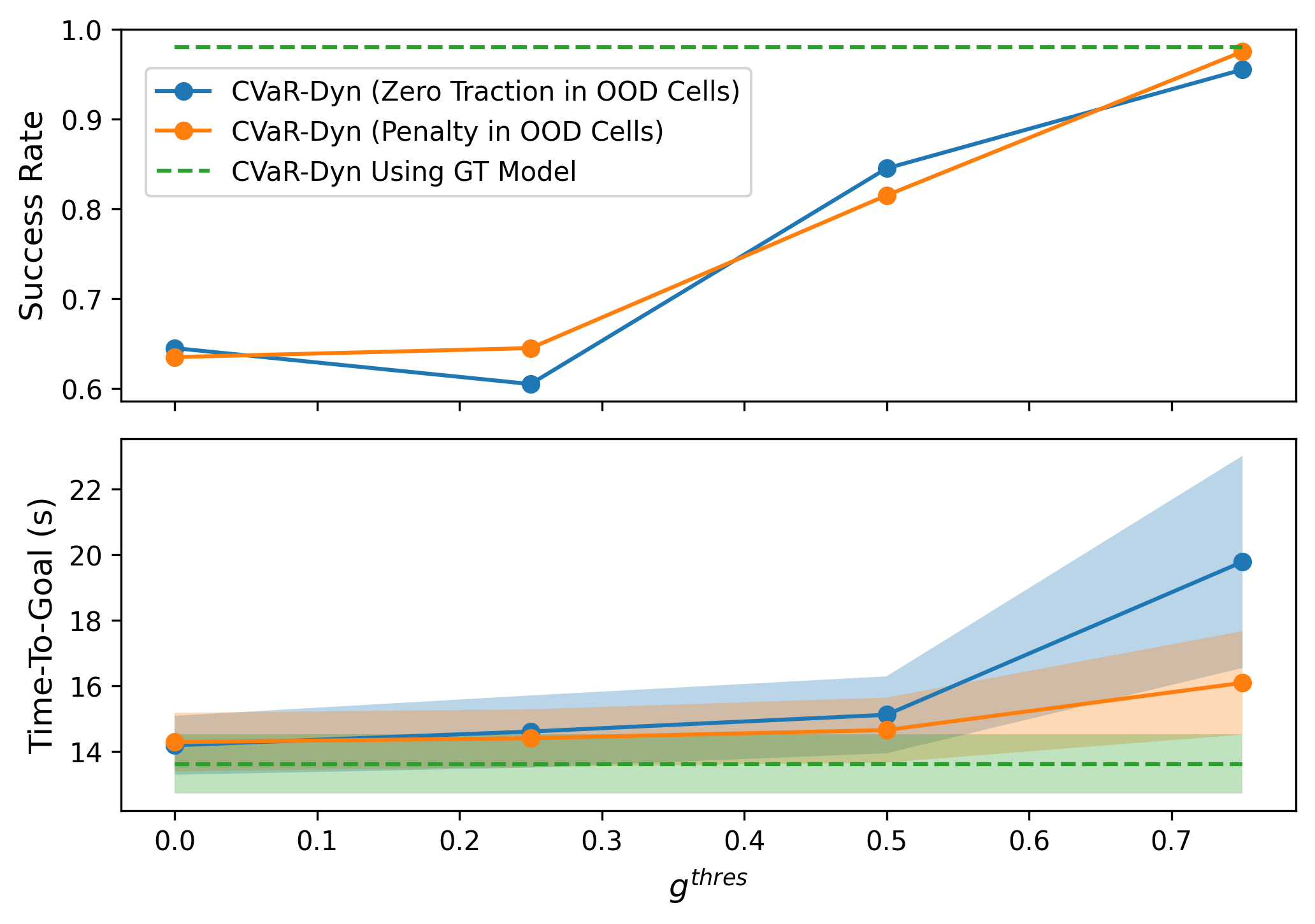}
	\caption{Navigation success rate improves thanks to the proposed OOD terrain detector when deploying a trained NN in a novel environment. As the focus is not on benchmarking different planning algorithms, we select \cvardyn{} with $\alpha=0.2$ and $g^{\text{thres}}=0$ as the baseline planner, but similar conclusions can be drawn using a different planner. The OOD terrains are handled by either assigning zero traction (blue) or imposing penalties (orange). The performance of the  planner that uses the ground truth (GT) traction is also shown to demonstrate the best performance. Overall, higher $g^{\text{thres}}$ improves the navigation success rate at the cost of higher time-to-goal, because there are more OOD terrain cells to avoid. However, auxiliary penalties for OOD terrains make it easier for a given planner to find solutions that lead to the goal. Notably, the average success rate when $g^{\text{thres}}=0.75$ approaches $1$, indicating that the learned traction model generalizes well to terrains with high confidence values in the test environment.
	}
	\label{fig:conf_score_benchmark} 
	\vspace*{-0.15in}
\end{figure}

In this section, we demonstrate the benefit of the proposed density-based confidence score~\eqref{eq:gmm_conf} for detecting terrains that lead to unreliable NN predictions. Because benchmarking different planning algorithms is not the main focus of this section, we use the proposed \cvardyn{} planner that has been shown to achieve higher success rates than \cvarcost{} and set a low risk tolerance $\alpha=0.2$ (experience has shown that similar results occur using other planners). In order to simulate training and testing environments, we leverage the data collected in two distinct forests, where the first one (visualized in Fig.~\ref{fig:data_collection}) is used to train the traction predictor, and the second one (whose semantic top-down view is shown in Fig.~\ref{fig:test_env}) is used to simulate the test environment. The traction values will be drawn from the test environment's empirical traction distribution learned by a separate NN as the proxy ground truth. 
Two specific start-goal pairs are selected in order to highlight the most challenging parts of the test environment with novel features. Each start-goal pair is repeated 10 times for each selected confidence threshold $g^{\text{thres}}$. We investigate two different approaches to prevent the planner from entering OOD terrains with low confidence: (1) assigning zero traction to OOD terrains, and (2) adding large penalties for states entering OOD terrains. Note that the mission is successful if each goal is reached within 30~s.

As shown in Fig.~\ref{fig:conf_score_benchmark}, the navigation success rate improves by up to $30\%$ as $g^{\text{thres}}$ increases, because the robot avoids regions where the network's prediction may be significantly different from the ground truth. Interestingly, using \cvardyn{} with additional penalties for states entering OOD terrains leads to better time-to-goal while retaining similar success rate. Intuitively, the auxiliary costs make it easier for the \cvardyn{} planner to find trajectories that move around the OOD terrains. Therefore, it is advantageous to combine the proposed cost formulations with auxiliary costs when domain knowledge is available in order to achieve both high success rate and fast navigation in practice.

\section{Conclusion \& Future Work}
This work proposed a probabilistic traversability model that is easy to train thanks to self-supervision that captures the full empirical distribution of the traction parameters in the unicycle dynamics. For navigation tasks in simulated environments, planning with the proposed risk-aware costs led to better performance than methods that assumed no slip or used expected traction. 
Furthermore, the learned traction model generalized better in novel environments by avoiding terrains that had low confidence scores based on the GMM-based density estimator. Lastly, using the proposed costs with auxiliary penalties for undesirable terrains, when such prior knowledge is available, can lead to improved performance.

Based on our results, the proposed \cvarcost{} planner achieved the best time-to-goals but suffered from poor sample efficiency and conservativeness. Therefore, two interesting research directions are to design a sample-efficient planner that optimizes the CVaR of the objective, and to investigate other risk metrics that are less conservative. Additionally, the proposed framework can be streamlined by replacing the GMM with a normalizing flow model that can be jointly trained. Lastly, extensive hardware experiments are needed to validate the proposed approaches in practice.

\section*{Acknowledgment}
Research was sponsored by ARL W911NF-21-2-0150 and by ONR grant N00014-18-1-2832.

\balance
\bibliographystyle{IEEEtran}
\bibliography{bibs}

\end{document}